\documentclass[lettersize, 10pt, conference]{ieeeconf}
\usepackage{mathptmx} % assumes new font selection scheme installed
\usepackage{times} % assumes new font selection scheme installed
\usepackage[dvipdfmx]{graphicx}
\usepackage{url}
\usepackage{graphicx}
\usepackage[figurename=Fig.\ ]{caption}
\usepackage{epsfig} % for postscript graphics files
\usepackage{cite}
\usepackage{color}
\usepackage{booktabs}  % 
\usepackage{amsmath}
\usepackage{amssymb}
\usepackage{indentfirst}
\usepackage{algorithm}
\usepackage{algorithmicx}
\usepackage{algpseudocode}
\usepackage{caption}
\usepackage{subcaption}
\usepackage{comment}
\usepackage{color,soul}
\usepackage{ifthen}
\usepackage[table,xcdraw]{xcolor}

\title{\LARGE \bf
Interactive Task Encoding System for Learning-from-Observation
}

\author{
Naoki Wake$^{1}$,
Atsushi Kanehira$^{1}$,
Kazuhiro Sasabuchi$^{1}$,
Jun Takamatsu$^{1}$,
and Katsushi Ikeuchi$^{1}$
\thanks{$^{1}$Applied Robotics Research, Microsoft, 
        Redmond, WA 98052, USA
        {\tt\small naoki.wake@microsoft.com}}%
}

\begin{document}
\bstctlcite{IEEEexample:BSTcontrol}
\maketitle
\thispagestyle{empty}
\pagestyle{empty}

\begin{abstract}
We present the Interactive Task Encoding System (ITES) for teaching robots to perform manipulative tasks. ITES is designed as an input system for the Learning-from-Observation (LfO) framework, which enables household robots to be programmed using few-shot human demonstrations without the need for coding. In contrast to previous LfO systems that rely solely on visual demonstrations, ITES leverages both verbal instructions and interaction to enhance recognition robustness, thus enabling \textit{multimodal LfO}. ITES identifies tasks from verbal instructions and extracts parameters from visual demonstrations. Meanwhile, the recognition result was reviewed by the user for interactive correction. Evaluations conducted on a real robot demonstrate the successful teaching of multiple operations for several scenarios, suggesting the usefulness of ITES for multimodal LfO. The source code is available at https://github.com/microsoft/symbolic-robot-teaching-interface.
\end{abstract}

\section{Introduction}
Household robots with manipulation capabilities are increasingly being considered as an alternative labor force in various settings including home environments~\cite{smarr2014domestic}. While typical robotic systems assume that a robot performs specific operations in a fixed environment, household systems need to provide the ability to adjust operations to fit the user's needs and environment. Learning-from-Observation (LfO) is a framework that aims to teach manipulative operations through human demonstrations without coding, making it a promising solution for programming household robots~\cite{ikeuchi1994toward}.

\begin{figure}[tb]
  \centering
  \includegraphics[width=0.42\textwidth]{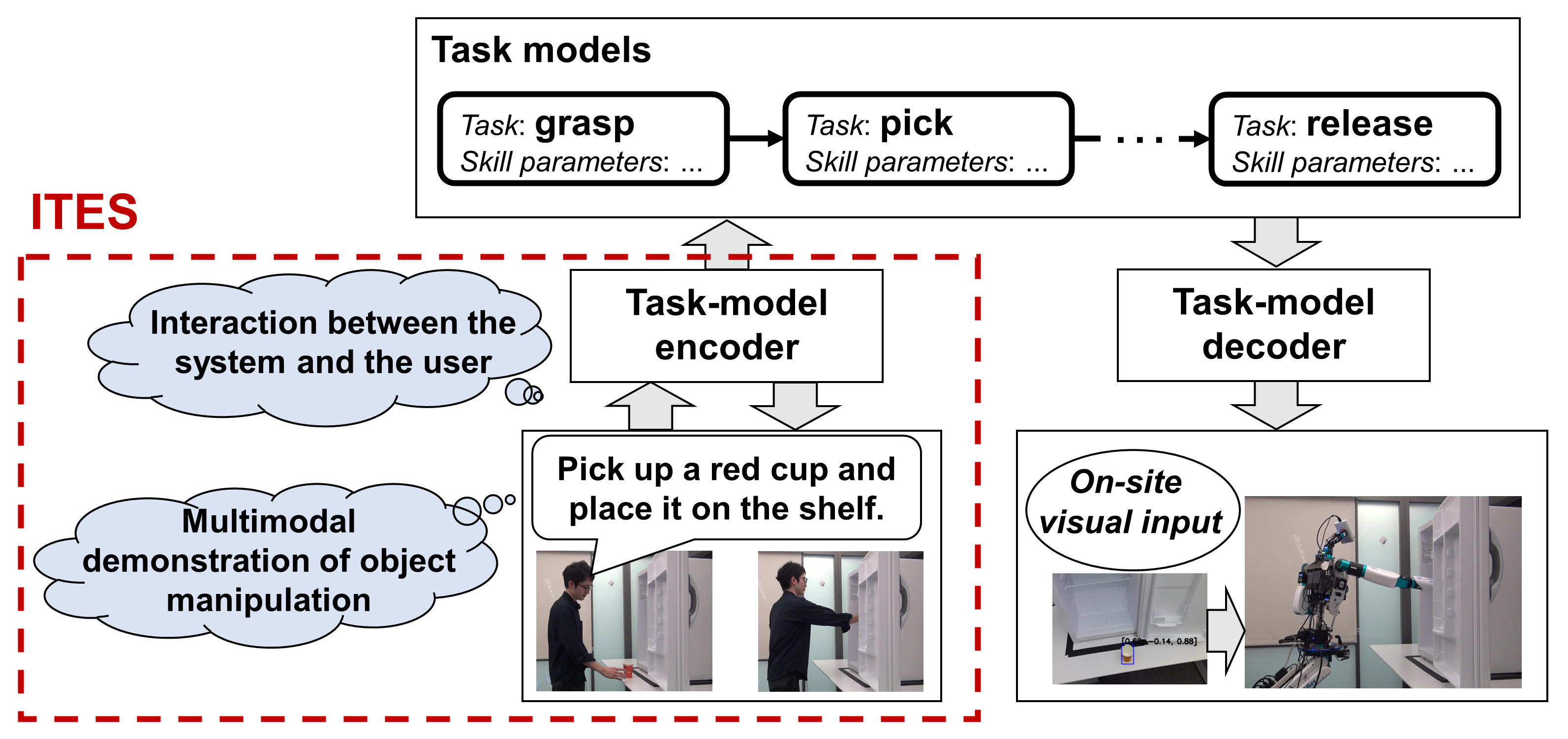}
  \caption{
  Overview of the robot teaching framework, so-called Learning-from-Observation (LfO). The red box indicated the Interactive Task Encoding System (ITES) for LfO. ITES encodes multimodal human demonstrations into a sequence of primitive robot actions, referred to as \textit{task models}.
  }
  \label{fig:title}
  % \vspace{-15pt}
\end{figure}

In LfO, a human demonstration is encoded into an intermediate representation of object manipulation, so-called a \textit{task model} (Fig.~\ref{fig:title}). The task model consists of a sequence of primitive robot actions, so-called \textit{tasks}, and parameters to achieve the tasks, so-called \textit{skill parameters}. Because the task model is an abstract representation of object operations, the encoded task model is theoretically applicable to multiple environments and arbitrary hardware.

Although studies have shown successful implementation of LfO systems in various settings, they have been primarily based on visual demonstration. For instance, vision-based LfO systems have been developed for part assembly~\cite{takamatsu2007recognizing, perez2017c, zhu2016inferring}, knot tying~\cite{takamatsu2006representation}, and dancing~\cite{ikeuchi2018describing, nakaoka2007learning, okamoto2014extraction}. On the other hand, studies have shown that human uses verbal instructions to make teaching more interactive and efficient~\cite{soto2007automatic, justice2008influence, lupyan2010making}. 
Inspired by the nature of language, we have developed an LfO system that utilizes both visual and verbal information, namely, \textit{multimodal LfO}~\cite{wake2020verbal, wake2020learning}. However, existing multimodal LfO systems have focused on robust visual recognition and information enrichment, and applications that take advantage of interactivity have not yet been proposed.

This paper aims to present a practical pipeline of the task-model encoder that employs interaction (Fig.~\ref{fig:title}). We refer to the encoder as the Interactive Task Encoding System (ITES). 
To enable users to teach at the task granularity level, we adopt a simple method of ``stop-and-go demonstration.'' In the stop-and-go demonstration, a user pauses the hand motions when tasks switch. At every pause, the demonstrator gives a verbal instruction for the next task before resuming the hand movements. Given the stop-and-go demonstration, ITES recognizes tasks and skill parameters and outputs a task model, with improved stability achieved through GUI-based interactive correction. The contributions of this paper include proposing a practical pipeline for multimodal LfO with interaction capability and demonstrating the system's effectiveness on a real robot with a dexterous hand.

\section{Related works}\label{section:related}
\subsection{Role of language in recognition}
Natural language has been shown to be useful in guiding vision systems. For example, we have previously shown that the use of verbal input can help a robot-vision system determine the timings and the location of object grasping and thus make the recognition robust~\cite{wake2020verbal}. In addition, verbal instructions include semantic constraints in manipulations that are not explicitly taught through visual demonstration~\cite{ikeuchi2021semantic}.

Language can be applied not only to vision-system guidance but also to interactive education. Recent research has proposed interactive systems that utilize human language for remembering users' previous inputs~\cite{brock2021personalization}, operations in new environments~\cite{ilyevsky2021talk}, and clarifying uncertain instructions~\cite{lanza2020abductive}. However, to our knowledge, no multimodal LfO system has been proposed that actively utilizes user interaction. This study proposes a system that implements user interaction to effectively utilize verbal instructions for task recognition.

\subsection{Robot task planning from language}
Starting with the work of the SHRDLU~\cite{winograd1972understanding}, the problem of task planning based on language instructions has been studied for decades.
Before the development of Natural language processing (NLP) technology, the dominant method of understanding sentences was by parsing verbs and syntax (e.g., ~\cite{thomas2012roboframenet, tellex2011understanding}). In recent years, the use of NLP technology has become more common, and systems have been proposed to analyze abstract instructions~\cite{ahn2022can, wake2023chatgpt} and complex instructions, including conditional branching~\cite{pramanick2020decomplex}. %For example, the authors in ~\cite{ahn2022can} propose a framework that decomposes abstract instruction into step-by-step procedures using large language models. 

Multimodal robot teaching has been studied in both end-to-end approaches based on large models~\cite{ahn2022can, ito2022integrated, Lynch-RSS-21, jang2021bc, stepputtis2020language, lynch2020language} and symbolic robot teaching approaches including multimodal LfO~\cite{Gopalan-RSS-20, venugopalan2015sequence, yanaokura2022multimodal, wake2020learning}. However, the previous LfO systems used a simple knowledge base that maps tasks to verbs in a single sentence, which limits the flexibility of input texts and the length of task sequences to be taught at once. In this paper, we aim to teach tasks of arbitrary length through more flexible sentences by utilizing a step-by-step teaching method and a task recognition model based on NLP.

\section{Teaching strategy for multimodal LfO}\label{section:multimodalLfO}
%The computation required for ITES is ``given a visual and speech demonstration, recognize the sequence of tasks and their temporal correspondence with the visual demonstration.'' 
%This section defines the inputs and outputs of the ITES and describes the difference between human demonstration and task execution granularity. We propose a teaching method to bridge this granularity difference. The computation required for ITES is to recognize the sequence of tasks and their temporal correspondence with visual and speech demonstrations.

%ITES requires recognizing the temporal correspondence between tasks and visual/speech demonstrations. This section defines ITES inputs/outputs and proposes a teaching method to bridge the granularity gap between human demonstration and task execution.
In the context of robot teaching, the granularity gap between human demonstrations and robot execution can be problematic. This section proposes a teaching method to bridge this gap.
\subsection{Unit of human demonstration and robot execution}
In this paper, we define that a unit of human demonstration starts with grasping an object followed by several manipulative tasks, and ends with releasing the object. We call this unit a grasp-manipulation-release (GMR) operation. We believe that various manipulative household operations can be achieved as a result of multiple GMR operations. For example, cleaning the table after a meal can be broken down into the GMR of a plate for clearing dishes and the GMR of a sponge for wiping the table. 
%; LfO can be applied to a wide range of household tasks by being able to learn the units of GMR. 
Note that we focused on single-armed operations and assumed that only one object is being manipulated in a GMR operation. 

%\subsection{Unit of robot execution}
On the other hand, the unit of robot execution, \textit{task}, can be finer than GMR operations. In a typical LfO, a task is defined as a transition of a target object’s state. For example, we have proposed a task set based on the types of motion constraints imposed on the object~\cite{ikeuchi2021semantic}. Consequently, a GMR operation is divided when the motion constraints to the object change. As an example, a GMR operation of ``picking up a cup and carrying it on the same table'' is divided into the tasks of picking up the object (PTG11) bringing it (PTG12), and placing it (PTG13) (symbols in parentheses are from ~\cite{ikeuchi2021semantic}). Table.~\ref{table:task-cohesion} shows examples of GMR operations and their task components. In this paper, we also include grasp and release in the task set because these actions involve a transition in the contact state between the robot hand and the object.

\begin{table}[tb]
	\centering
	\caption{Examples of GMR operations (modified from ~\cite{yanaokura2022multimodal})}
	\renewcommand\tabcolsep{4pt}
	\begin{tabular}{lc}
		\toprule  %
		GMR operation & Explanation \\
		\midrule
        Grasp-PTG11-PTG12-PTG13-Release & pick, bring, and place something \\
        Grasp-PTG31-Release & slide something open \\
        Grasp-PTG33-Release & slide something close \\
        Grasp-PTG51-Release & rotate something open \\
        Grasp-PTG53-Release & rotate something close \\
		\bottomrule  %
	\end{tabular}
	\label{table:task-cohesion}
	% \vspace{-15pt}
\end{table}

\begin{figure}[tb]
  \centering
  \includegraphics[width=0.45\textwidth]{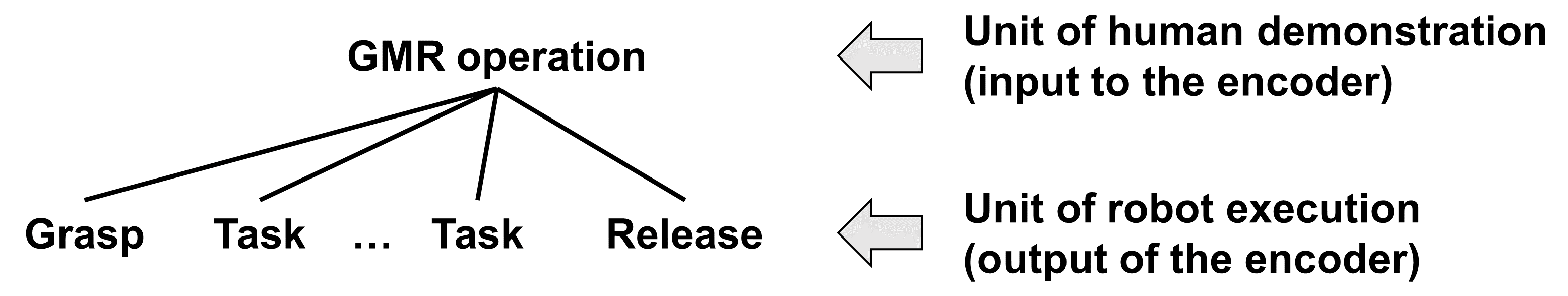}
  \caption{
  Manipulative operations with different granularity. 
  }
  \label{fig:GMR-taskcohesion-task}
  % \vspace{-15pt}
\end{figure}

\subsection{Stop-and-go teaching to resolve granularity differences}\label{section:stop-and-go}
We designed a teaching method that allows users to effectively teach at the granularity of tasks. First, to enable users to teach multiple tasks in a GMR operation at once, we adopted a stop-and-go demonstration technique. This method involves pausing the hand movements when tasks switch in order to inform the system of the task transition. For instance, in a pick-and-place operation, the demonstrator stops the hand motion when approaching the object, grasping it, lifting it off the table, carrying it above the table, placing it on the table, releasing it, and moving the hand to a home position.

Additionally, we adopted a teaching method that alternates between visual demonstration and verbal instruction, rather than simultaneously. This is because simultaneous teaching is considered a form of dual tasking that requires a higher cognitive load for the user. Such teaching methods may not be suitable for inexperienced users, especially for the elderly~\cite{verhaeghen2003aging}. To clearly convey the relationship between verbal and visual demonstrations, the demonstrator provides step-by-step verbal instruction for the next task at each pause, rather than giving verbal instructions all at once before or after visual demonstrations.
%In order to explicitly teach the correspondence between verbal and visual demonstrations, the demonstrator gives step-by-step verbal instruction for the next task at the timing of each hand stop, rather than giving verbal instructions all at once before or after visual demonstrations.

\section{Implementations of ITES}\label{section:implementation}
\begin{figure}[tb]
  \centering
  \includegraphics[width=0.45\textwidth]{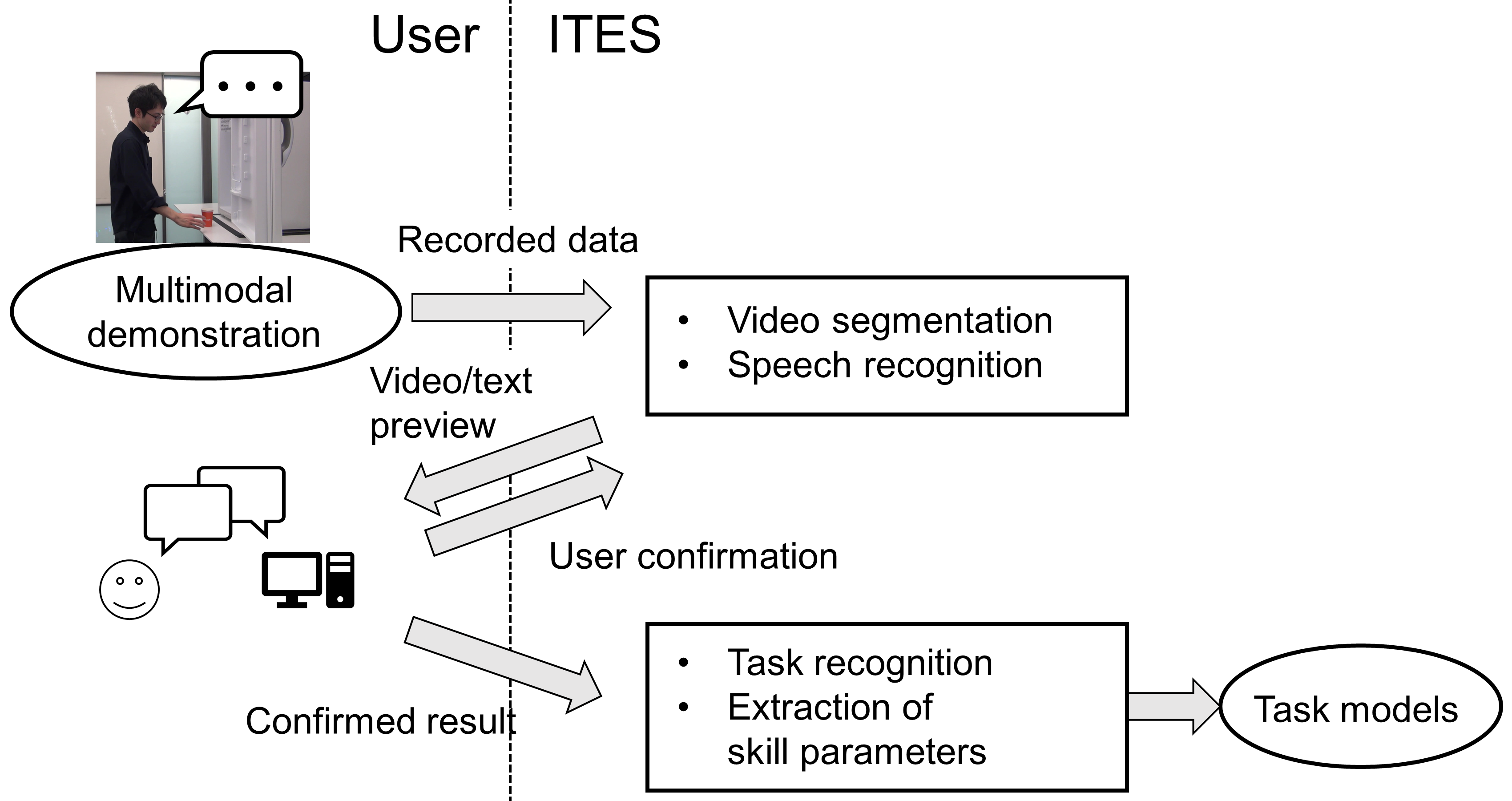}
  \caption{
  The pipeline of ITES, which involves interaction with users. 
  }
  \label{fig:pipeline}
  % \vspace{-15pt}
\end{figure}

%This section explains the implementation of ITES (Fig.~\ref{fig:pipeline}). ITES consists of two stages before and after an interaction with the user. In the first stage, ITES detects segmenting timings based on the stopping motion and splits the video and speech based on the timings. The split speech is transcribed using a third-party speech recognizer. The result of segmentation and speech recognition was previewed so that the user modify the result if necessary. In the later stage, a task sequence is recognized from the verbal input using an NLP-based recognizer, and skill parameters are extracted for each task to compile a task model.
This section describes the implementation of ITES (Fig.~\ref{fig:pipeline}). ITES comprises two stages, one before and one after user interaction. In the initial stage, ITES detects times when a manipulating hand stops and segments the video and speech accordingly. The speech is transcribed using a third-party speech recognizer, and the result of segmentation and speech recognition are previewed for user modification if necessary. In the later stage, an NLP-based recognizer identifies the task sequence from the verbal input, and skill parameters are extracted for each task to compile a task model.

\subsection{System input}
We use an Azure Kinect sensor~\cite{k4a} to record RGB-D images and a speech input during the demonstration. The sensor is placed in positions that could capture the entire demonstration. The image resolution is 1280x720, and the nominal sampling rate for the video and speech is 30 Hz and 48000 Hz, respectively. Noise in the speech was suppressed using noise filters to improve speech recognition~\cite{wake2019enhancing}.

\subsection{Video segmentation and speech recognition}\label{segmentation}
\begin{figure}[tb]
  \centering
  \includegraphics[width=0.40\textwidth]{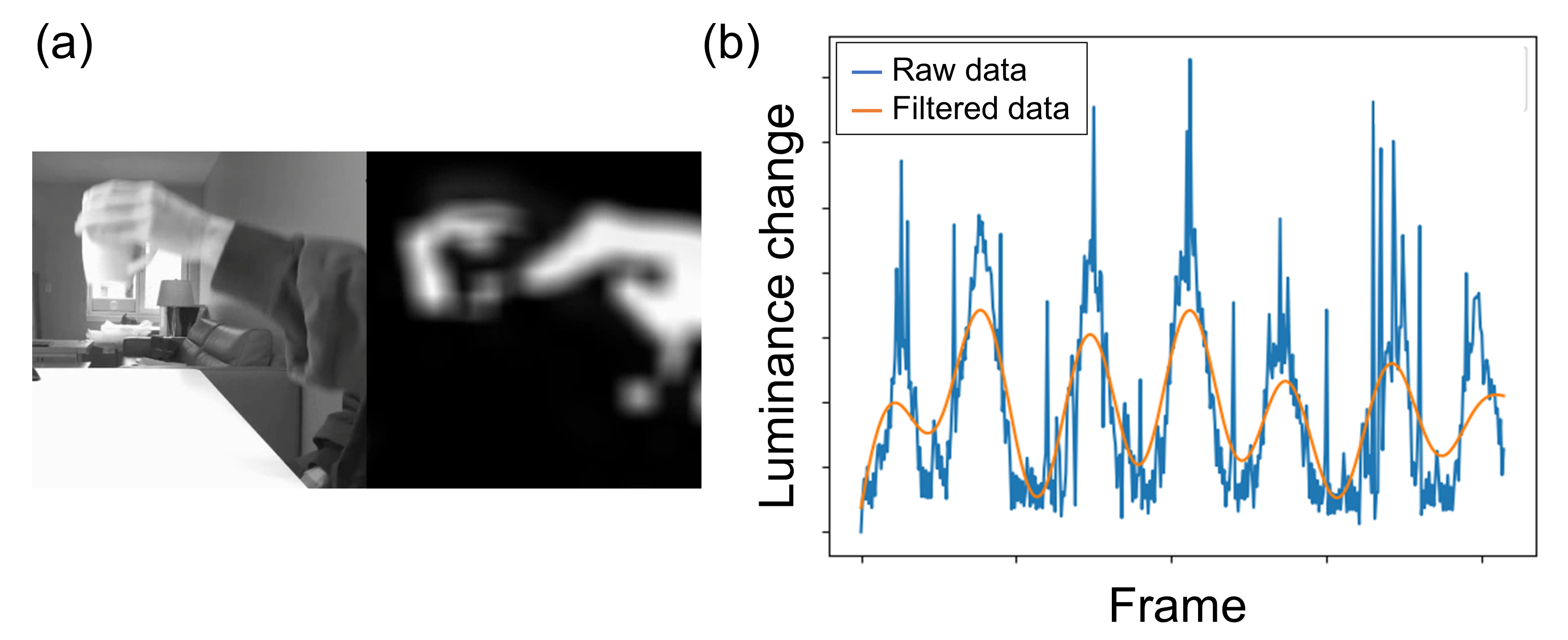}
  \caption{
  (a) An example of the pixel-wise absolute difference between adjacent luminance images (right panel). In the video, the hand is picking up the cup. (b) Time series of the luminance change for a stop-and-go demonstration.
  }
  \label{fig:segmentation}
  % \vspace{-15pt}
\end{figure}

Given a stop-and-go demonstration, ITES segments the video and audio at the times when a manipulating hand stopped. To detect the times, we characterized the intensity of hand motions based on the changes in luminance~\cite{ikeuchi1992towards}. For this calculation, RGB images are converted to YUV images, and the Y channel is extracted as the luminance. 
The luminance images are spatially filtered using a moving average of 50x50 window, and the pixel-wise absolute difference is calculated between adjacent frames (Fig.~\ref{fig:segmentation}(a)). The mean of the difference is taken as the luminance change at each time. After removing outliers and applying a low-pass filter of 0.5 Hz, the local minima of the time series are extracted as the stop timings (Fig.~\ref{fig:segmentation}(b)). After the confirmation of the split timings (in Sec.~\ref{preview_feature}), the RGB-D video and the audio are segmented based on the detected timings. The split speech is transcribed using a third-party cloud speech recognition service~\cite{azure-speech-recognition} and previewed to the user for additional confirmation.

%\subsection{Previewing for the user modification}\label{preview_feature}
%Since the speech content and the correspondence between video and audio are critical to task recognition and the extraction of skill parameters, we implement the feature to ask for users' modifications on the computation result of video splitting and speech recognition. After the luminance-based video splitting, the user is prompted for each split video. The user can ignore videos unrelated to the teaching (e.g., movements before and after a GMR operation) and merge over-split videos through a button-based GUI (Fig.~\ref{fig:feedback}). Note that the user gives the demonstration again when the under-segmentation occurred.% instead of manually segmenting the video through the GUI.
\subsection{Previewing for the user modification}\label{preview_feature}
\begin{figure}[tb]
  \centering
  \includegraphics[width=0.42\textwidth]{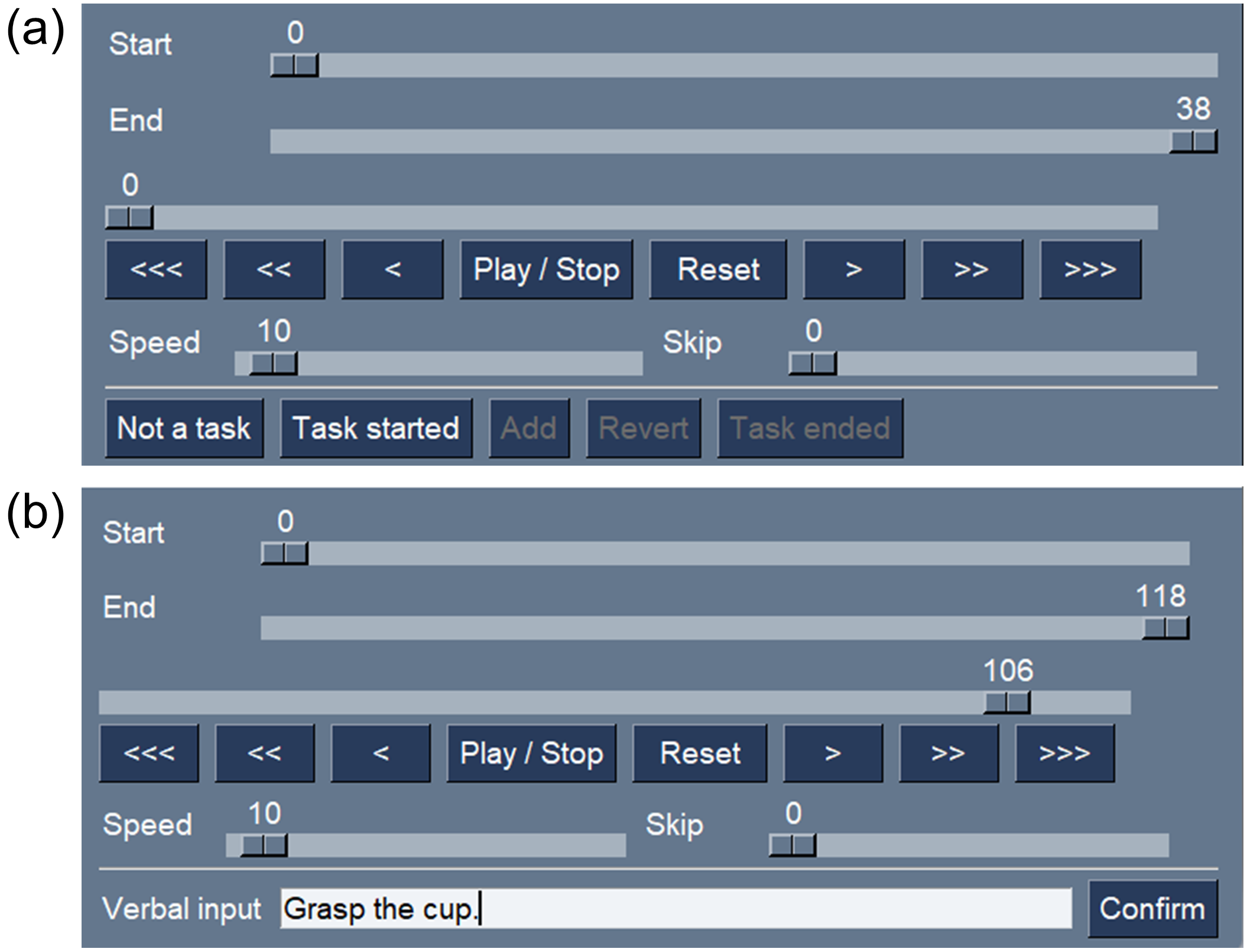}
  \caption{
  User interfaces for checking the computation result of video segmentation and speech recognition. (a) An interface to ignore and merge videos that were segmented by ITES. (b) An interface to correct speech recognition results. 
  }
  \label{fig:feedback}
  % \vspace{-15pt}
\end{figure}

Since the speech content and the correspondence between video and audio are critical to task recognition and the extraction of skill parameters, we implemented a feature that allows users to modify the computation result of video segmentation and speech recognition. After the luminance-based video segmentation, the user is prompted for each split video. The user can ignore videos unrelated to the teaching (e.g., movements before and after a GMR operation) and merge over-split videos through a button-based GUI (Fig.~\ref{fig:feedback} (a)). Note that the user gives the demonstration again when the under-segmentation occurred.

After confirming the video segmentation, the video and audio are split again, and the resulting audio segments are transcribed into text using a speech recognition service~\cite{azure-speech-recognition}. The transcribed text is then displayed to the user in order so that the user can modify through a GUI (Fig.~\ref{fig:feedback} (b)).

\subsection{Task recognition}
After the interactive correction, each task is recognized based on the transcribed texts using a language-based recognition model. To train the model, we manually annotated an existing video dataset of preparing breakfast~\cite{saudabayev2018human}. We chose the cooking domain because cooking needs the use of a variety of foods and tools with manipulation. We labeled the video with task labels and prepared the video dataset of a single task using a third-party video annotation tool. The dataset contained 12 task classes with a total of 1340 videos. Table~\ref{table:annotation} shows the task classes and the number of data.

\begin{figure}[tb]
  \centering
  \includegraphics[width=0.26\textwidth]{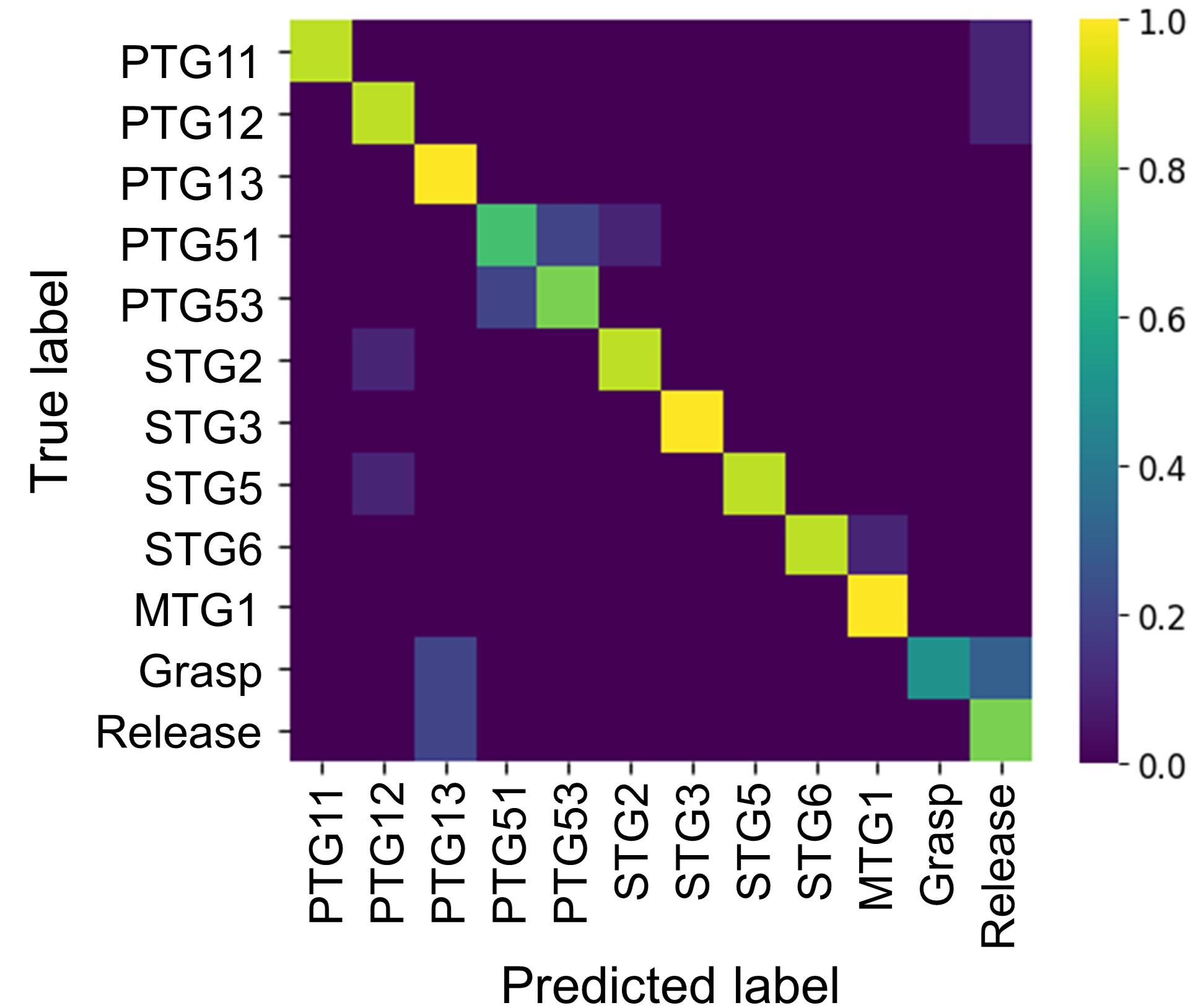}
  \caption{
  Confusion matrix of task recognition from texts.
  }
  \label{fig:confusionmatrix}
  % \vspace{-15pt}
\end{figure}

We collected motion instructions for the egocentric videos of each task using a crowdsourcing service, called Amazon Mechanical Turk. Specifically, 100 instruction sentences were collected for each task. Then, we trained a recognition model to associate each instruction with its corresponding task. To this end, we prepared a random forest model trained on top of a fixed BERT model~\cite{devlin2018bert}. %Note that we collected the data using egocentric videos because the demonstrator is assumed to give demonstrations while teaching, rather than while observing the demonstration from a third-person perspective. 
We collected egocentric videos to match the perspective of the demonstrator during teaching, rather than from a third-person perspective.
Table~\ref{table:sentence_examples} shows examples of sentences collected by different cloud workers. We observed variations in the verbs and nouns that appear in the instructions. The confusion matrix of task recognition is shown in Fig.~\ref{fig:confusionmatrix}, where 10\% of the sentences were used for testing. We conducted ten-fold cross-validation and obtained an average performance of 83\%. This result suggests that the system can robustly recognize tasks from natural verbal instructions with variations.

\begin{table}[]
\caption{Statics of the labeled tasks. Task symbols in ~\cite{ikeuchi2021semantic} are indicated in parentheses.}
\begin{tabular}{|l|c|c|}
\cline{1-3}
\hline
\multicolumn{1}{|c|}{\cellcolor[HTML]{E7E6E6}Task label} &
\multicolumn{1}{c|}{\cellcolor[HTML]{E7E6E6}Count} &
\multicolumn{1}{c|}{\cellcolor[HTML]{E7E6E6}Mean length (S.E.) (sec)}
\\ \cline{1-3}
Picking (PTG11)  & 122 & 0.63 (0.03) \\ \cline{1-3}
Bringing (PTG12) & 108  & 1.46 (0.20)  \\ \cline{1-3}
Placing (PTG13)  & 113 & 0.73 (0.05)  \\ \cline{1-3}
Rotating\_hinge\_to\_open (PTG51) & 38  & 1.14 (0.14) \\ \cline{1-3}
Rotating\_hinge\_to\_close (PTG53) & 32 & 0.75 (0.11)   \\ \cline{1-3}
Wiping  (STG2)   & 30 & 5.59 (1.42) \\ \cline{1-3}
Peeling (STG3)   & 24  & 3.55 (0.90)  \\ \cline{1-3}
Pouring (STG5)   & 42  & 2.26 (0.54)  \\ \cline{1-3}
Holding (STG6)   & 239 & 3.50 (0.32)  \\ \cline{1-3}
Cutting (MTG1)   & 22  & 4.52 (0.75)  \\ \cline{1-3}
Grasping         & 464 & 0.56 (0.02)  \\ \cline{1-3}
Releasing        & 392  & 0.37 (0.01)  \\ \cline{1-3}
\end{tabular}
\label{table:annotation}
\end{table}

\begin{table}[]
    \centering
    \caption{Examples of sentences collected using a crowd-sourcing service.}
    \begin{tabular}{lc}%{|l|}
        %\hline
        \toprule  %
         A video of opening a refrigerator door \\ 
        %\hline
        \midrule
            ``Open the refrigerator door.'' \\ 
            ``Pull to open the fridge door.''\\
            ``Grab the refrigerator handle and pull it to open the door.''\\
        %\hline
        \midrule
        %\hline
        A video of washing a place with a sponge \\
        %\hline
        \midrule
            ``Wipe the dish with the sponge in an anti-clockwise direction.''\\
            ``Clean the plate with the sponge in your hand.''\\ 
            ``Wipe the plate with the sponge.''\\
        %\hline
        \midrule
        %\hline
        A video of pouring water from a kettle\\ 
        %\hline
        \midrule
            ``Pour water from the kettle into the mug.''\\
            ``Pour water from the pitcher into the cup.''\\
            ``Empty the water from the kettle into a cup.''\\
        %\hline
        \bottomrule
        \end{tabular}
    \label{table:sentence_examples}
\end{table}

\subsection{Skill parameter extraction}\label{skillparameters}
Every task requires skill parameters for robots to decode. ~\cite{wake2020verbal, wake2020learning}. Once a task is determined, processes called daemon run to extract the parameters by analyzing the corresponding video. Here we briefly explain the examples of skill parameters and computation for extracting the parameters.

\subsubsection{The name and the 3D positions of the target object}
The name of the target object is used to help recognize other skill parameters. We predefined a set of object name and assumed that users would specify object names verbally. In the current ITES, an object name is extracted using a third-party language parser~\cite{manning2014stanford}. The 3D positions of the grasped object is calculated from the RGB-D images.
\subsubsection{Hand laterality}
Hand laterality is crucial information that relates to other skill parameters, such as grasp type and approach direction to the target object. To extract hand laterality, the video of grasping the object is analyzed. Assuming a constant object location, the 2D location of the object is extracted from the first RGB image using an object detector. In the last frame, the 2D locations of both hands are detected using a hand detector, and the hand used for manipulation is identified as the hand closer to the object.

\subsubsection{Grasp type}
Grasp type is critical for successful task execution of tasks followed by the grasp~\cite{saito2022task}. ITES detects the grasp type using an image classifier model. The image of the manipulating hand at the last frame of the grasping video is input into the model, and the grasp type is determined. The likelihood of grasp types associated with the object name is also considered to improve the accuracy of the classification~\cite{wake2021object, wake2020grasp}.

\subsubsection{Hand positions and hand trajectory}
\begin{figure}[tb]
  \centering
  \includegraphics[width=0.34\textwidth]{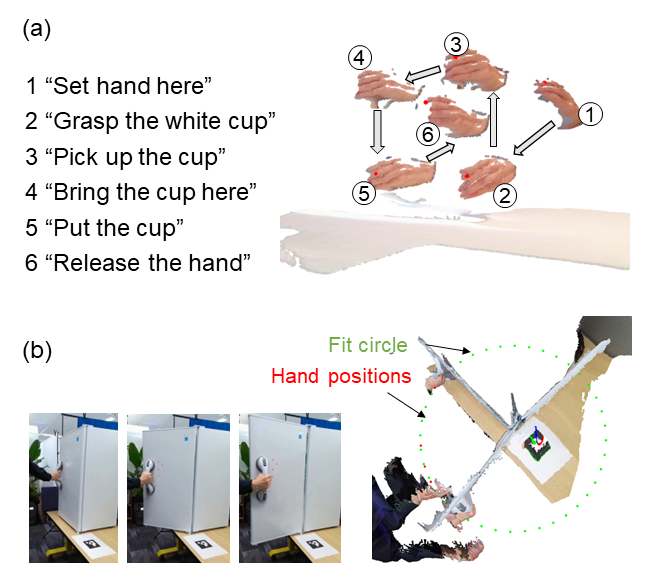}
  \caption{
  (a) Extracted hand positions and (b) the estimation of the rotating\_hinge task.
  }
  \label{fig:skillparameter}
  % \vspace{-15pt}
\end{figure}

The demonstrator's hand motion is crucial for successful task completion and avoiding collisions with the environment, as well as for manipulating articulated objects such as doors or shelves. ITES extracts 3D hand positions during the demonstration using the 2D hand detector and depth images (Fig. ~\ref{fig:skillparameter}(a)). For rotating\_hinge tasks, the hand trajectory is parameterized by applying circular fitting to the hand positions during the corresponding video (Fig. ~\ref{fig:skillparameter}(b)).

\subsubsection{Human pose}
Human pose contains implicit knowledge for achieving tasks efficiently~\cite{sasabuchi2020task}. Following our previously proposed task model design~\cite{wake2020learning}, we encode the human arm postures at the start and end of each task video. The 3D poses of the demonstrator are estimated using a third-party 3D pose estimator~\cite{mediapipe}, and each of the body parts (upper/lower arms of left and right arms) is encoded into spatially digitized 26-point directions on the unit sphere.

%\subsubsection{3D Object positions}
%The 3D position of the grasped object is calculated from the RGB-D image taken at the start of grasping. During robot execution, our LfO system recalculates the object position using an on-site visual input (see Fig.\ref{fig:title}), assuming that the object position may slightly differ from that of the demonstration. Positional information is represented relative to an AR marker placed in the environment.

\section{Experiments}~\label{section:experiments}
We tested the proposed ITES based on the performance of multimodal LfO. To this end, we prepared a multimodal LfO system by integrating ITES into a task-model decoder that we had previously implemented. The control policies of robots were trained using reinforcement learning~\cite{saito2022task, takamatsu2022} on top of our in-house simulator~\cite{sasabuchi2023}. Since ITES is designed to be applicable to a wide range of home environments, the LfO system was qualitatively examined from two aspects: (1) if the system applies to a wide variety of GMR operations by combining tasks, (2) if the system provides flexibility for users to adjust a GMR operation in different scenes.

To check the (1) system applicability, we tested three GMR operations commonly observed in household situations: ``pick-carry-place a box,''``throw away a cup,'' and ``open a fridge.'' To check the (2) system flexibility, we considered a case of a pick-carry-place operation that accompanies multiple \textit{bring (PTG12)} tasks to avoid obstacles. we tested the system with a humanoid robot~\cite{nextage} and a dexterous robot hand with four fingers~\cite{shadowrobot_2022}.
 
\subsection{GMR of pick-bring-place an object}
\begin{figure}[tb]
  \centering
  \includegraphics[width=0.35\textwidth]{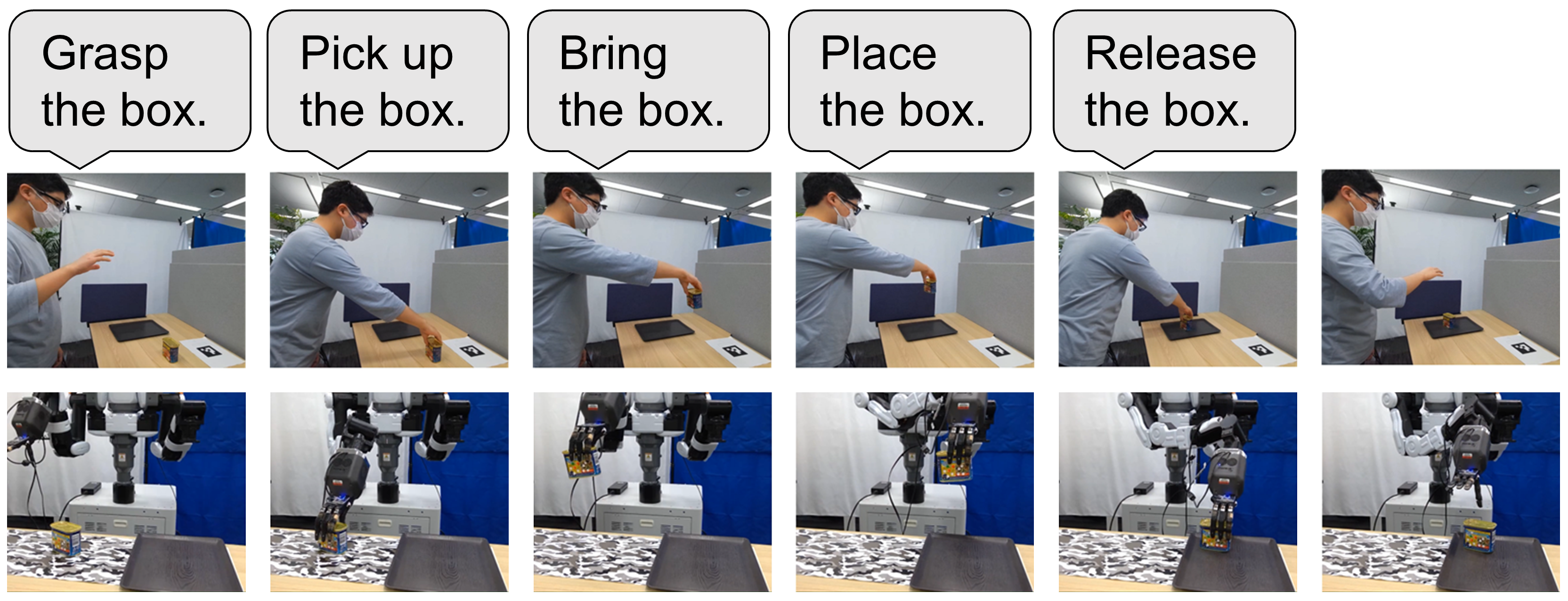}
  \caption{Results of multimodal LfO for pick-carry-place a box.
  }
  \label{fig:GMR_pickplace}
  % \vspace{-15pt}
\end{figure}

Fig.~\ref{fig:GMR_pickplace} shows the overview of teaching a ``pick, carry, and place a box'' operation. This GMR operation consists of \textit{grasp}, \textit{pick (PTG11)}, \textit{bring from one location to another (PTG12)}, \textit{place (PTG13)}, and \textit{release} tasks.
The verbal input and the visual demonstration are shown at the top of the figure. %The two robots
The result shows that the demonstration was successfully executed by the robot. To check the system flexibility, we tested the ``pick-carry-place'' operation with a box placed on a shelf (Fig.~\ref{fig:GMR_throw_open}(a)). For this environment, several waypoints are required to avoid collision with the shelf, thus user instructed multiple \textit{bring (PTG12)} tasks in-between the \textit{pick (PTG11)} and \textit{place (PTG13)} tasks. The result shows that the multiple bring tasks can be executed by the robot, suggesting that the system has the flexibility to allow users to adjust GMR operations to different scenes.

\subsection{GMR of throwing away a cup and opening a door}
\begin{figure}[tb]
  \centering
  \includegraphics[width=0.48
  \textwidth]{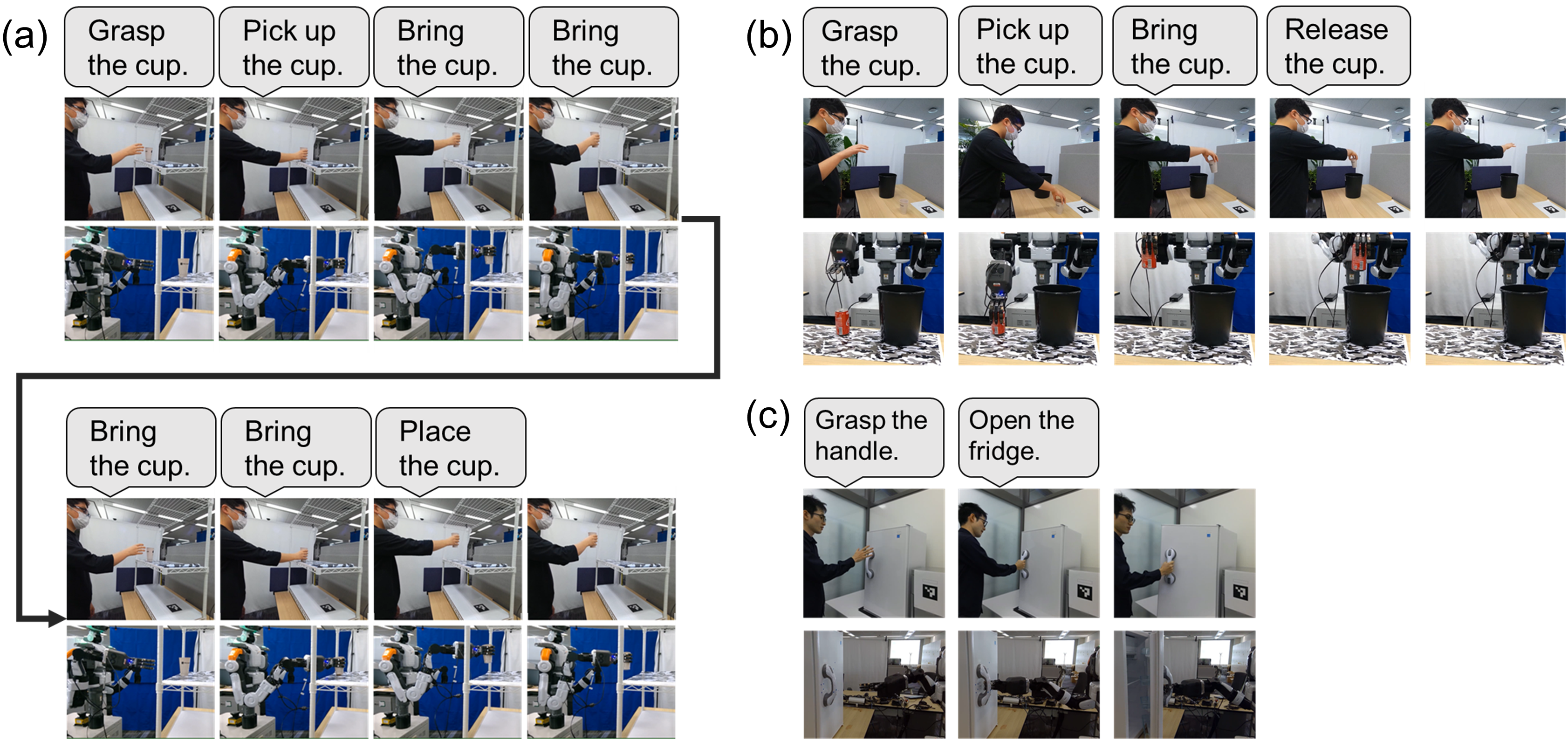}
  \caption{Results of multimodal LfO for the GMR operation of (a) pick-carry-place a cup with multiple bringing tasks in-between, (b) throwing away a cup, and (c) opening a door.
  }
  \label{fig:GMR_throw_open}
  % \vspace{-15pt}
\end{figure}

%We tested if the system applies to a variety of GMR operations by teaching additional GMR operations of
We tested the applicability of the system to a variety of GMR operations by teaching additional tasks, including the operations of throwing a cup away (Fig.~\ref{fig:GMR_throw_open}(b)) and opening a door (Fig.~\ref{fig:GMR_throw_open}(c)). The throwing operation consists of \textit{grasping}, \textit{picking (PTG11)}, \textit{bringing from one location to another (PTG12)}, and \textit{releasing} tasks. The opening operation consists of \textit{grasping}, \textit{opening to a certain range (PTG5)}, and \textit{releasing} tasks. Those operations were completed by the robot, suggesting that the proposed LfO system can operate a variety of operations by composing the tasks.

\subsection{Failure cases of task-model recognition}
During the experiment, we observed several failure cases. One is the case where the recorded visual demonstration was not segmented correctly due to inefficient hand-stop time. Additionally, we observed several cases when skill-parameter extraction failed due to the incorrect recognition of hands, objects, grasp types, and human poses. Such misrecognition was typically caused by occlusion. For example, an occluded human hand caused the failure of extracting hand position and trajectory. In those cases, the user needed to discard the demonstration and start over.

\section{Discussion}~\label{section:discussion}
In this paper, we proposed ITES, a pipeline of the task-model encoder for multimodal LfO. By assuming alternating stop-and-go visual demonstration and step-by-step verbal instruction, GMR operations can be taught at the granularity of tasks while taking correspondence between visual and verbal inputs. Additionally, ITES featured an interaction function in order to allow the user to modify the recognition results (Fig. ~\ref{fig:pipeline}). Experiments tested the applicability and flexibility of ITES for various GMR operations in different scenes. As a result, a robot successfully executed the taught GMR operations. Although not included in this paper, we have additionally confirmed that the encoded GMR operations were operated by another robot with different degrees of freedom (results are shown in~\cite{takamatsu2022}). These results suggest the usefulness of ITES. %The contributions of this paper include proposing a practical pipeline for multimodal LfO with interaction capability and demonstrating the system's effectiveness on a real robot.

The design philosophy of LfO is similar to low-code/no-code software development (LCSD), which enables non-programmers to participate in software development with minimal coding through a visual platform~\cite{rokis2022challenges,sahay2020supporting}. Likewise, LfO allows non-programmers to program household robots without coding. % by leveraging their knowledge of housework. 
However, robot manipulation requires many skill parameters that are difficult to represent in languages, such as arm postures and hand trajectories; some of them are based on common sense and may not be consciously apparent to the demonstrator~\cite{ikeuchi2023applying}. Thus, specifying such parameters solely on a visual platform can be time-consuming. ITES addresses this challenge by extracting the parameters from visual demonstration. Our approach extends LCSD to handle multimodal input.%Therefore, a robot programming system for non-programmers would be efficient by extending LCSD to handle multimodal input.

The method of providing multimodal input is not unique. As described in Sec.~\ref{section:stop-and-go}, the present ITES employed a teaching method that alternates between visual demonstration and verbal instruction. This approach has the advantage of explicitly teaching the correspondence between verbal and visual inputs while limiting the cognitive load on the user. 
However, some users may prefer simultaneous teaching to reduce teaching time, while others may prefer to give all verbal instructions at the beginning or the end. 
A further user study is needed to determine which teaching method is appropriate from a usability perspective.

\section*{Acknowledgments}
We thank Dr. Sakiko Yamamoto (Ochanomizu University), Dr. Etsuko Saito (Ochanomizu University), and Dr. Midori Otake (Tokyo Gakugei University) for their help and advice in creating the video dataset of tasks.

\bibliographystyle{IEEEtran}
\bibliography{bib}

\end{document}